\documentclass{article}
\usepackage{spconf,amsmath,graphicx}
\usepackage{multicol}
\usepackage{lipsum}
\usepackage[english]{babel}

\usepackage{fancyhdr}
\fancypagestyle{plain}{%
  \fancyhf{}%
  \fancyfoot[C]{\footnotesize Page \1\ of \pageref{4}}%
}

\title{A RADIOMICS APPROACH TO TRAUMATIC BRAIN INJURY PREDICTION IN CT SCANS}
\name{Ezequiel de la Rosa$^{1,2}$, Diana M. Sima$^{1}$, Thijs  Vande  Vyvere$^{1}$, Jan S. Kirschke$^{2}$, Bjoern Menze$^{2}$ }
\address{$^{1}$ico\textbf{metrix}, Leuven, Belgium\\ $^{2}$Department of Computer Science, Technical University of Munich, Germany}

\begin{document}


\maketitle

\begin{abstract}

\end{abstract}
Computer Tomography (CT) is the gold standard technique for brain damage evaluation after acute Traumatic Brain Injury (TBI). It allows identification of most lesion types and determines the need of surgical or alternative therapeutic procedures. However, the traditional approach for lesion classification is restricted to visual image inspection. In this work, we characterize and predict TBI lesions by using CT-derived radiomics descriptors. Relevant $shape$, $intensity$ and $texture$ biomarkers characterizing the different lesions are isolated and a lesion predictive model is built by using Partial Least Squares. On a dataset containing 155 scans (105 train, 50 test) the methodology achieved 89.7\% accuracy over the unseen data. When a model was build using only texture features, a 88.2\% accuracy was obtained. Our results suggest that selected radiomics descriptors could play a key role in brain injury prediction. Besides, the proposed methodology is close to reproduce radiologists decision making. These results open new possibilities for radiomics-inspired brain lesion detection, segmentation and prediction.

\section{Introduction}
\label{sec:intro}
Traumatic Brain Injury (TBI) is a complex disease process that  encompasses a whole spectrum of different pathologies. In the acute phase after injury, non-contrast Computed Tomography (CT) is the most commonly used imaging modality. It can detect most abnormalities, but is especially important for identifying the presence of large extra-or intra-axial space-occupying lesions (i.e, subdural hematomas, epidural hematomas or contusions).  For instance, extra-axial hematomas that cause mass effect (i.e, basal cistern compression and midline shift) may need urgent neurosurgical evacuation. On the other hand, contusions may require a non-surgical treatment approach. In this regards, detection and classification of these lesions is of paramount importance in the medical decision-making process \cite{parizel2005new}.\\
$Radiomics$, ``the conversion
of digital medical images into
mineable high-dimensional data" \cite{gillies2015radiomics} takes advantage of image analysis techniques for describing underlying physiopathology in medical scans. As a result, a descriptive feature vector is obtained and subsequent interpretation by computer driven techniques is performed. For dealing with the high-dimensional space, dimensionality reduction methods turns crucial. In the last few years, the multivariate Partial Least Squares (PLS) method for analyzing radiomics-derived descriptors has been explored. The technique is extensively used in the $OMICS$ field, since it is  suitable  for  problems where the number of features is larger than the number of samples.
\\
In this work, we aim to characterize and predict TBI lesions using CT-derived descriptors. The main contributions of this work are $i$) identification of distinctive radiomic biomarkers characterizing brain lesions, $ii$) fitting of PLS models that accurately predict lesion classes and $iii$) exploring whether $texture$-based models outperform $shape$ and $intensity$-based ones for the desired task.

\section{Materials and Methods }
\label{sec:format}

\subsection{Database}
In this work, data from the CENTER-TBI study (www.center-tbi.com, NCT02210221) coming from more than 50 academic and non-academic centres in 20 countries was used. CT volumes were obtained from several scanners from all major manufacturers, including General Electric, Siemens, Philips and Toshiba. Images were acquired following several acquisition and reconstruction parameters, with variations on slice thickness, pixel spacing and scanner settings, among others.\\
A sub-cohort of 3D volumes was randomly chosen, including three type of brain mass lesions: $i)$ epidural hematoma (class 1), $ii)$ acute subdural hematoma (class 2) and $iii)$ contusion (class 3). 105 scans (train set) were used for data analysis, characterization, and model fitting. Besides, we included 50 extra scans (test set) for validating our models. The training (test) set contains 170 (69) annotated lesions, with 49 (20) in class \#1, 54 (20) in class \#2 and 67 (29) in class \#3. Volume's slices were delineated by three trained operators, supervised by an experienced neuroradiologist. Hyperdensities were annotated using 3D Slicer (https://www.slicer.org/).

\subsection{Pre-processing}
With the aim of extracting robust biomarkers insensitive to scan acquisition and reconstruction parameters, all scans were resliced into an homogenous voxel dimension of [1x1x1 mm$^{3}$]. As demonstrated in \cite{shafiq2017intrinsic}, voxel resampling is a strongly recommended step for obtaining reproducible radiomic descriptors.

\subsection{Radiomics Feature Extraction}
For each lesion present in the scan we extracted radiomic features using the ground truth segmentations. When a volume contained more than one type of lesion, separate descriptors for each of them were considered. With this aim, the open-source $PyRadiomics$ \cite{van2017computational} library was used and 105 features from the original volumes were obtained grouped as follows: $i)$ Shape (13), $ii$) First Order Statistics (FOS, 18), $iii)$ Gray-level Co-occurrence Matrix (GLCM, 23), $iv$) Gray level Difference Matrix (GLDM, 14), $v$) Gray-level Run Length Matrix (GLRLM, 16), $vi$) Gray-level Size Zone matrix (GLSZM, 16)
    and $vii$) Neighborhood Gray-Tone Difference Matrix (NGDM, 5).\\
Afterwards we built the so called feature \textbf{X} ($N\times 105$) and responses \textbf{Y} ($N\times 3$) matrices ($N$ being the number of lesions in the training set).   

\subsection{Data Analysis: Partial-Least-Squares}
With the aim of characterizing brain lesions using radiomics and generating a predictive TBI model, we used the multivariate PLS technique. PLS discriminant analysis (PLS-DA) is a flexible tool used for descriptive and predictive modelling, as well as for discriminant feature selection. The algorithm incorporates dimensionality reduction with discriminant analysis for high-dimensional data interpretaion. In a nutshell, the algorithm involves two steps: i) PLS latent variables (LVs) construction and ii) predictive model building \cite{lee2018partial}. The LVs are computed as linear combinations of the independent \textbf{X} variables, \textbf{XW}, where the loading weights in \textbf{W} are chosen in such a way that the corresponding LVs have maximal covariance with the responses in the \textbf{Y} matrix. For multiclass problems, \textbf{Y} is a dummy matrix encoding the class membership information. After building the model, the responses of unknown class data can be predicted from their independent variables \cite{aliakbarzadeh2016classification}.

\subsubsection{Feature Selection}
For finding informative and distinctive markers allowing classes discrimination, the Variables Importance on Prediction (VIP) criterion was used. VIP scores help in detecting and ranking those features contributing in the model fitting. For each variable, the VIP score is equal to the accumulated weights (\textbf{w}) over all selected LV's. For feature selection it has been widely suggested to retain those features with VIP's greater than the unity \cite{aliakbarzadeh2016classification}. 

\subsubsection{Model Fitting}
In this work, firstly a model using all the considered features was fitted. Secondly, by using the above explained VIP criterion, the model was retrained with the retained features only. Model fitting was performed over the training set in a 10-fold cross-validation strategy by changing the number of LV's and by assessing the classification error rate. The model that minimized the classification error was preferred. Afterwards, PLS models were validated by predicting the unseen test samples.  

\subsection{Experiments}
With the aim of assessing the feature-class effect for predicting the lesions, PLS models were assessed under different feature combinations. 

\begin{itemize}
    \item {\textbf{Experiment 1}} All features considered.
    \item {\textbf{Experiment 2}} Feature selection over all feature classes.
    \item {\textbf{Experiment 3}} Feature selection considering only $Shape$ descriptors.
    \item {\textbf{Experiment 4}} Feature selection considering only $Shape$ and $FOS$ descriptors.
    \item {\textbf{Experiment 5}} Feature selection considering only $texture$ (GLCM, GLDM, GLRLM, GLSZM, NGDM) descriptors .
\end{itemize}

Selected experiments were chosen on the basis of comparing radiologist's $observable$ descriptors (shape and FOS) and assessing their importance on lesion labelling. On the other hand, we explore as well the models behaviour under the inclusion of more complex computational descriptors.

\subsection{Performance and statistics}
The most relevant per-class descriptors in terms of VIP scores were assessed by means of the non-parametric Kruskal-Wallis test. When p-values exhibited statistical significance, the Dunn’s test for multiple comparisons was performed. The Benjamini \& Hochberg procedure for controlling the false discovery rate was applied. Two-tailed tests with a 0.05 significance level were used. For addressing mass lesion classifiers using PLS-DA, accuracy, sensitivity and specificity were used as performance metrics.

\begin{table*}[h]
\caption{Fitted models summary and number (\%) of selected features.  }
\begin{tabular}{p{.3cm}p{.4cm}p{.3cm}p{1.5cm}p{1.37cm}p{1.27cm}p{1.25cm}p{1.25cm}p{1.25cm}p{1.25cm}p{1.25cm}p{1.25cm}}
    &                 &                & \multicolumn{9}{c}{\textbf{Features}}                                                                                        \\
    \cline{4-12}
    
    & \multicolumn{2}{c}{\textbf{Model}} & \textbf{Considered} & \multicolumn{8}{c}{\textbf{Selected}}                                                                           \\
    \cline{2-3}
    \cline{5-12}
\textbf{Exp} &\textbf{ER}           & \textbf{LV}          & \textbf{Total}      & \textbf{Total}        & \textbf{Shape}       & \textbf{FOS}        & \textbf{GLCM}       & \textbf{GLDM}       & \textbf{GLRLM}      & \textbf{GLSZM}      & \textbf{NGDM}      \\
\hline
\#1 & 0.12            & 5              & 105        & 105 (100) & 13 (100)  & 18 (100) & 23 (100) & 14 (100) & 16 (100) & 16 (100) & 5 (100) \\
\#2 & 0.10            & 7              & 105        & 40 (38.1) & 12 (92.3) & 8 (44.4) & 8 (34.7) & 1 (7)    & 7 (43.7) & 4 (25)   & 0 (0)   \\
\#3 & 0.25            & 4              & 13         & 4 (30.7)  & 4 (30.7)  &            &            &            &            &            &           \\
\#4 & 0.15            & 7              & 31         & 11 (35.5) & 6 (46.1)  & 5 (27.7) &            &            &            &            &           \\
\#5 & 0.13            & 15             & 74         & 17 (22.9) &             &            & 6 (26.1) & 1 (7)    & 6 (37.5) & 4 (25)   & 0 (0) \\
\hline
\end{tabular}
Note: Exp: Experiment; ER: Error Rate.
\label{table:models}
\end{table*}

\begin{figure*}[!hbtp]
\includegraphics[width=17.7cm]{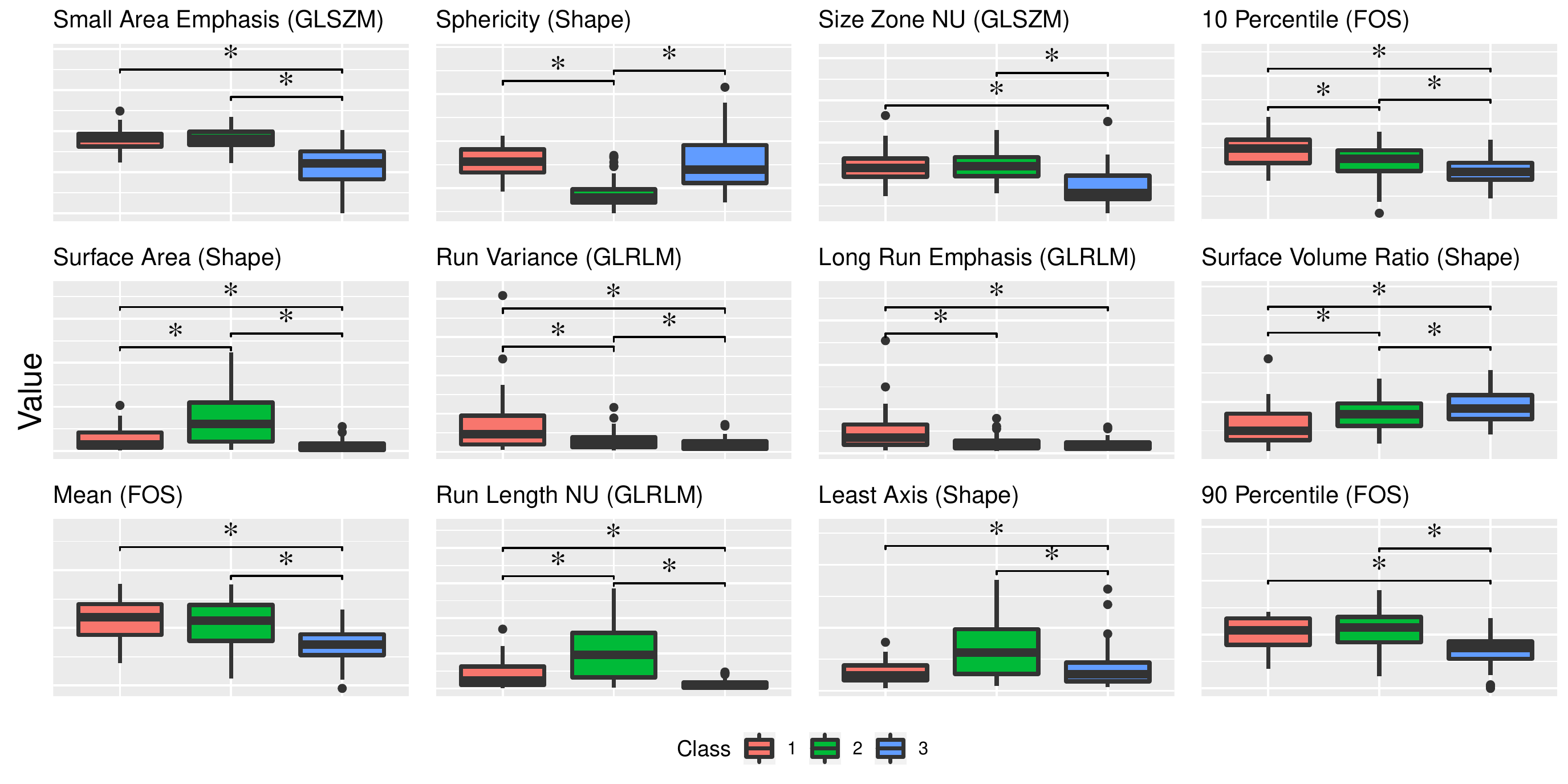}
\caption{Common selected features boxplots. Classes 1, 2 and 3 belong to $epidural$ $hematoma$, $acute$ $subdural$ $hematoma$ and $contusion$. NU: Non-uniformity.  *: significant p-value.}
\label{fig:boxplots}
\end{figure*}

\section{RESULTS}
\label{sec:RESULTS}
A summary of the predictive models fitness obtained for each experiment is shown in Table \ref{table:models}. As expected, the model where feature selection was conducted considering all descriptors (Exp \#2) obtained the lowest error-rate. Besides, it outperformed the model where no feature selection was applied (Exp \#1), suggesting that non-explanatory features have been discarded.\\
The worst performing model in terms of errors was the one fitted in Exp \#3 (only shape information considered). It is observable that these features where frequently chosen in the different experiments (12 out of 13 shape features are also used by the best model, namely Exp \#2). Despite being frequently chosen, our results suggest that shape descriptors are not sufficient for distinctively characterizing the lesion classes. However, when $FOS$ features were added to the $shape$ model, the error-rate considerably decreased. It is worth mentioning that these are $observable$ features for radiologists, having a valuable meaning for lesion labelling.\\
When texture features were only used for fitting the model (Exp \#5), a low error-rate was obtained. Among all feature classes, GLSZM and GLCM features were preferred. The NGDM features have never been selected, suggesting non informative descriptors for the considered task. \\
Boxplots of the shared retained features among the different models are shown in Fig. \ref{fig:boxplots}. It is possible to appreciate that all these features exhibited statistical significance among some or all groups. Considering radiological interpretable descriptors,  $Sphericity$ showed to be a distinctive one. The results are supported by lesion morphology configurations, namely that subdural ones are less spherical than epidural and contusion ones \cite{parizel2005new}. On the other hand, $Surface$ $Areas$ were greater for subdural hematoma when compared against the other lesions.\\
Table \ref{table:perf} shows achieved classification performance on the unseen data. The best performance was obtained for model \#2, which improved by $\sim 7 \%$ on accuracy the model without feature selection. When shape information was used for fitting the model, a low 72.1 \% accuracy was obtained. Inclusion of $FOS$ descriptors  were able to considerably improve the model, by performing $\sim 10 \%$ better on accuracy terms. For Exp \#5, $texture$ features were able to almost reproduce the best performing model by achieving 88.2\% accuracy. This result suggests texture features encode relevant information for differentiating among TBI lesions. The importance of this latter model relies on its independence from shape descriptors. Since texture analysis can be performed patch-wise, the technique suggests potential for lesion detection, segmentation and classification.   

\begin{table}[h]
\caption{Classification performances.}
\begin{tabular}{p{.5cm}p{.5cm}p{.69cm}p{.69cm}p{.69cm}p{.69cm}p{.69cm}p{.69cm}}
           &          & \multicolumn{3}{c}{\textbf{Se}}       & \multicolumn{3}{c}{\textbf{Sp}}       \\
           \cline{3-8}
\textbf{Exp} &\textbf{Acc}  & \textbf{C \#1} &\textbf{C \#2}  & \textbf{C \#3} & C \textbf{\#1} &\textbf{C \#2}  &\textbf{C \#3}  \\
\hline
\#1        & 82.3     & 85        & 89.5      & 75.9      & 89.6      & 89.8      & 94.9      \\
\textbf{\#2}         & \textbf{89.7}     & \textbf{80}        &\textbf{94.7}       &\textbf{93.1}       &\textbf{100}        &\textbf{89.8}       &\textbf{94.9}       \\
\#3        & 72.1     & 60        & 94.7      & 65.5      & 95.8      & 73.5      & 89.7      \\
\#4        & 83.8     & 75        & 89.5      & 86.2      & 97.9      & 85.7      & 92.3      \\
\#5        & 88.2     & 85        & 94.7      & 86.2      & 95.8      & 93.9      & 92.3 \\
\hline
\end{tabular}
Note: Exp: Experiment; Acc: Accuracy (\%); Se: Sensitivity (\%); Sp: Specificity (\%); C\#: Classes.
\label{table:perf}
\end{table}


\section{DISCUSSION AND CONCLUSIONS}
\label{sec:DISCUSSION}
In this work several PLS models using radiomic CT features were compared. Three different brain lesion types were considered and characterized by means of different descriptors. We isolated several radiomarkers that may play a key role in discrimination of TBI lesions, all behaving differentially among classes (p-val significant). We devised, as well, an automatic method for classifying the lesions. When using selected descriptors chosen over all feature types, a 89.9\% classification accuracy over the test set was achieved. Our results suggest that radiomic features are very close to reproduce radiologists decision making. When only using shape descriptors, it was found that $Sphericity$, $Surface$ and the $Surface$ $to$ $Volume$ $Ratio$ play an important role. However, shape features were not enough to accurately discriminate the lesion types. Since intensity and histogram derived metrics represent crucial information in medical imaging, we included $FOS$ and $Shape$ features in our models. Even by achieving a consistent improvement ($\sim 10 \%$ greater accuracy), these features were not able to reproduce the obtained results with the whole database.    
An important result of this work regards the potential and capability of texture descriptors for characterizing the lesions. We were able to build a model using only texture information which predicts the test set with 88.2\% accuracy, behaving almost as the model that included all features. This result is relevant since it suggests texture CT information might help in lesion detection, multi-class segmentation and classification. Besides, it suggests the existence of non-observable data structures  helping in discriminating TBI damage. Potentially, the developed models could be used for performing automatic quality control of segmented and predicted lesions. It is important to keep in mind that radiological inspired markers might be preferred for clinical applications when conducting these tasks.

\section{acknowledgements}
This work was funded by the H2020 Marie-Curie ITN TRABIT (765148) project. Data used for this work came from CENTER-TBI (FP7-COOPERATION-2013-602150). We thank Charlotte Timmermans, Nathan Vanalken and Thijs Vande Vyvere for conducting the manual segmentations. 

\bibliographystyle{IEEEbib}


%
%
%

\vfill
\pagebreak


\end{document}